\def\year{2022}\relax
\documentclass[letterpaper]{article} 
\usepackage{aaai22}  
\usepackage{times}  
\usepackage{helvet}  
\usepackage{courier}  
\usepackage[hyphens]{url}  
\usepackage{graphicx} 
\usepackage{natbib}  
\usepackage{caption} 
\usepackage{hyperref}
\DeclareCaptionStyle{ruled}{labelfont=normalfont,labelsep=colon,strut=off} 
\usepackage{algorithm}
\usepackage{algorithmic}
\usepackage{amsmath}  
\usepackage{amsfonts}

\usepackage{bbm}

\usepackage{newfloat}
\usepackage{listings}
\floatstyle{ruled}
\newfloat{listing}{tb}{lst}{}
\floatname{listing}{Listing}

\usepackage{pdfpages}
\usepackage{bm}
\usepackage{float}
\usepackage{subcaption}
\usepackage{rotating}
\usepackage{booktabs,threeparttable}
\usepackage{multirow}
\usepackage{amsthm}

\title{TransBoost: A Boosting-Tree Kernel Transfer Learning Algorithm for Improving Financial Inclusion} 

\title{TransBoost: A Boosting-Tree Kernel Transfer Learning Algorithm for Improving Financial Inclusion}
\author {
    Yiheng Sun,\textsuperscript{\rm 1}
    Tian Lu,\textsuperscript{\rm 2}\footnotemark[1]
    Cong Wang,\textsuperscript{\rm 3}\footnote{corresponding author}
    Yuan Li,\footnote{This project was done while the author was at Google.}
     Huaiyu Fu,\textsuperscript{\rm 4}
    Jingran Dong,\textsuperscript{\rm 1}
    Yunjie Xu\textsuperscript{\rm 4}
}
\affiliations {
    \textsuperscript{\rm 1} Tencent Weixin Group\\
    \textsuperscript{\rm 2} Heinz College, Carnegie Mellon University\\
    \textsuperscript{\rm 3} Guanghua School of Management, Peking University\\
    \textsuperscript{\rm 4} School of Management, Fudan University\\
    elisun@tencent.com, lutiansteven@gmail.com, wangcong@gsm.pku.edu.cn,  Liyuanlp@google.com, hyfu20@fudan.edu.cn, jingrandong@tencent.com, yunjiexu@fudan.edu.cn 
}

\begin{document}
\maketitle
\begin{abstract}
\begin{quote}
The prosperity of mobile and financial technologies has bred and expanded various kinds of financial products to a broader scope of people, which contributes to  financial inclusion. It brings non-trivial social benefits of diminishing financial inequality. However, the technical challenges in individual financial risk evaluation exacerbated by the unforeseen user characteristic distribution and limited credit history of new users, as well as the inexperience of newly-entered companies in handling complex data and obtaining accurate labels, impede further promotion of financial inclusion. To tackle these challenges, this paper develops a novel transfer learning algorithm (i.e., \textit{TransBoost}) that combines the merits of tree-based models and kernel methods\footnote{Our implementation can be found \href{https://github.com/yihengsun/TransBoost}{here}.}. The TransBoost is designed with a parallel tree structure and efficient weights updating mechanism with theoretical guarantee, which enables it to excel in tackling real-world data with high dimensional features and sparsity in $O(n)$ time complexity. We conduct extensive experiments on two public datasets and a unique large-scale dataset from Tencent Mobile Payment. The results show that the TransBoost outperforms other state-of-the-art benchmark transfer learning algorithms in terms of prediction accuracy with superior efficiency, demonstrate stronger robustness to data sparsity, and provide meaningful model interpretation. Besides, given a financial risk level, the TransBoost enables financial service providers to serve the largest number of users including those who would otherwise be excluded by other algorithms. 
That is, the TransBoost improves financial inclusion.    

\end{quote}
\end{abstract}

\section{Introduction}

The rapid growth of mobile and financial technologies has bred various kinds of financial products such as microloan, consumer debt, and internet insurance \citep{teja2017indonesian}. Many companies have joined the boat and explored ways of offering novel financial products \citep{musto2015personalized}. These emerging financial products have dramatically expanded financial services to a broader range of people, better satisfying their financial needs \citep{guild2017fintech}. It greatly advocates \textit{financial inclusion}. Realizing financial inclusion has significant social benefits. Approximately two billion people do not have a basic bank account till 2017 \citep{worldbank2018}. Advancing financial inclusion can help reduce poverty by helping people invest in the future, smooth their consumption, and manage income shocks and financial risks, and thus increase sustainability of the entire society \citep{dev2006financial}. In the highly competitive financial markets, financial service providers are also incentivized to pursue more users. The consumer pool expansion is critical to improve their commercial sustainability \citep{demirgucc2017financial}. 

However, attenting to the pool of underserved users causes challenges in financial risk evaluation. The first challenge concerns the users who have limited credit history records, resulting in the lack of useful information for model training. This problem is severer in developing countries or regions that do not have mature social credit system. Another challenge is that new users might demonstrate different characteristic distributions from the existing ones 
 \citep{bassem2012social}. The patterns learned from the existing users may have limited generalizability and lead to poor predictive performance for new news. This problem is more serious for new financial service providers who are inexperienced in obtaining sufficient and accurate financial risk labels for model training and in handling complex large-scale user data.      

Transfer learning techniques emerge as useful tools for improving financial risk evaluation. Transfer learning deals with how systems can quickly adapt themselves to new situations, tasks, and environments by transferring knowledge from a related task learned \cite{yang2020transfer}. By leveraging transfer learning and the user datasets with financial risk labels (which usually come from users of existing products or data of similar contexts), financial practitioners can improve financial risk assessment accuracy for the nascent products for financial inclusion.

State-of-the-art transfer learning algorithms include kernel-based algorithms such as Kernel Mean Matching (KMM) \citep{huang2006correcting}, JDA \citep{long2013transfer}, and deep learning algorithms such as DAAN \cite{yu2019transfer}, DDAN \cite{wang2020transfer}, and DSAN \cite{zhu2020deep}. Although the kernel-based algorithms have sound theoretical properties,  these algorithms are usually computationally expensive,  typically at least $O(n\sqrt{n})$ time \cite{Rudi2018FALKON}.
The deep learning algorithms have succeeded in visual, audio and natural language processing tasks. However, these algorithms lack model interpretability \citep{jung2020overview}. 

Another challenge in financial risk evaluation is the sparsity of the large-scale dataset. Missing values and outliers are prevalent in real-world applications. Although some of the above-mentioned techniques can tackle empty value and outlier issues, most of them are costly and ineffective in handling complex business settings with highly dimensional features. The shortcomings lead to a low adoption rate of these algorithms in financial industry. By contrast, \textit{tree}-based algorithms have shown desired properties in  \textit{effectiveness} and \textit{efficiency} for financial risk evaluation. Especially, tree-based algorithms could outperform deep learning algorithms when being applied to structured dataset without underlying temporal or spatial interactions among features \citep{chen2016xgboost}. Besides, because tree-based algorithms grow in a binary split approach, they are naturally \textit{robust to sparse and noisy input} \citep{friedman2001greedy}. Finally, due to rule-based model structure, tree-based algorithms are usually more \textit{interpretable} than deep learning algorithms and even linear algorithms \citep{lundberg2020local}. 

We propose the \textit{TransBoost}, a novel transfer learning algorithm that combines the merits of tree-based models and kernel methods. Through a specially designed parallel boosting-tree framework, the algorithm can train the transfer learning classifier in only $O(n)$ time while outperforming state-of-the-art algorithms in prediction accuracy. We have proved that the algorithm is theoretically a generalization of KMM framework with boosting trees being a more flexible kernel method.

The contributions of this study are multi-fold:
\begin{enumerate}
\item To the best of our knowledge, we are the first to demonstrate the value of transfer learning for financial inclusion. We showcase how practitioners can leverage (open-source) existing dataset with labels to learn individual financial risk for a new domain, and thus expand financial products and services to a broader range of people.
\item Given that state-of-the-art transfer learning algorithms are either computationally expensive or lack of flexibility, we take advantage of tree-based and kernel methods, and develop an innovative boosting-tree transfer learning algorithm as a more flexible kernel trick for KMM. Theoretically and empirically, we show it is an efficient algorithm that yields KMM weights in $O(n)$ time.  
\item Our proposed boosting-tree kernel algorithm not only outperforms other algorithms in terms of prediction accuracy, but also is robust in handling noisy and high-dimensional data with superior efficiency, and yields meaningful model interpretation. This is non-trivial for practice because data sparsity, noise, and lack of interpretability have restrained many advanced algorithms from being applied for financial risk assessment in real-world business.  
\end{enumerate}

\section{Related Work}

\subsection{Financial Risk Evaluation and Financial Inclusion}
Scholars have made great strides in improving financial risk evaluation accuracy from data and algorithm perspectives. \citet{soto2021peer} did a complete summary. 

It is important for modern financial companies to pursue financial inclusion. Financial inclusion refers to the delivery of financial services at affordable costs to vast sections of disadvantaged and low-income groups \citep{dev2006financial}. Academic studies, government white books, and commercial reports have amply documented the importance of financial inclusion. \citet{karlan2010expanding} and \citet{cull92financial} summarized the benefits of financial inclusion and \citet{klapper2016achieving} reviewed how financial inclusion can help achieve the Sustainable Development Goals. Recently, financial companies have collected data from nontraditional sources to help assess the creditworthiness of individual users and small business owners of "thin-file" \citep{oskarsdottir2020credit}. However, this approach heavily relies on the availability of user data, which is always difficult and costly to obtain for most financial companies. 

\subsection{Transfer Learning}

A vital task for transfer learning is to minimize the distribution discrepancy across domains. Various transfer learning approaches have been proposed to fully leverage the known information, which can be categorized into two types: instance weighting and feature representation. Instance weighting methods estimate and assign different weights to known instances to mimic the target domain. For example, KMM approach estimates the weights via minimizing the mean across the source and target domains in a reproducing kernel Hilbert space (RKHS) \cite{huang2006correcting}. Tradaboost follows the design principle of AdaBoost ensemble method to iteratively update the weights of training data \cite{dai2007boosting}. Feature representation methods seek to find a common feature representation between the source and target domains to achieve knowledge transfer. Such common representation is usually carried out by mapping the source and target domain to another space, including transfer component analysis (TCA) \cite{pan2010domain}, joint distribution adaptation (JDA) \cite{long2013transfer}, balanced distribution adaptation (BDA) \cite{wang2017balanced}, etc., or aligning one to the other, e.g., Correlation Alignment (CORAL) \cite{sun2016return}. 
Some of these approaches have been proposed to solve complex problems in natural language processing as well as computer vision \cite{zhu2020deep,wang2020transfer}. 

\subsection{Gradient Boosting Decision Tree (GBDT)}

GBDT \citep{friedman2001greedy} is a widely used ensemble tree model framework. There are some successful implementations of GBDT in academia and industry, including XGBoost \citep{chen2016xgboost}, LightGBM \citep{ke2017lightgbm} and CATBoost \citep{dorogush2018catboost}. In brief, GBDT is a model with \textit{M} additive regression trees as its output. The model is trained iteratively. In each iteration, the tree is optimized by the first-order (GBDT) or second-order (XGBoost) Taylor approximation of the regularized objective function. 
Formally, given a dataset $D=\{(x, y)\}$, the model is defined as 
$$
h(x)=\sum_{k}^{M} T_{k}\left(x, \Theta_{k}\right),
$$
where $h(x)$ is GBDT and $\mathrm{M}$ is the total number of trees. $T_{k}$ is the tree of the \textit{k}-th iteration and $\Theta_{k}$ is the parameter. For the output of every tree, $T_{k}\left(x, \Theta_{k}\right)=$ $\{q(x),w(q(x))\}$, where $q(x)$ is the tree structure function mapping instance $x$ to a certain leaf $j$ with $w(j)$ as its weight.

\section{TransBoost Learning Algorithm}
A common transfer learning scenario in industry involves a large-sized labeled source domain from a matured product and a limited-sized labeled target domain from a newly launched product. The target of the TransBoost algorithm is to leverage the instances from matured product to improve model performance on target domain.
Formally, assuming that $\mathcal{D}=\left\{\mathcal{D}^{S}, \mathcal{D}^{T}\right\}$ is a dataset of source domain and target domain. $\mathcal{D}^{S}=\left\{(x, y) \mid\left(x_{i}^{S}, y_{i}^{S}\right) \in P^{S}, i=1, \ldots, N_{S}\right\}$ denotes $n_{S}$ instances from source domain and $
\mathcal{D}^{T}=\left\{(x, y) \mid\left(x_{j}^{T}, y_{j}^{T}\right) \in P^{T}, j=1, \ldots, N_{T}\right\}
$ denotes $n_{T}$ instances from target domain with $\mathrm{m}$ features $ ( \left.x \in \mathbb{R}^{m}, y \in \mathbb{R}\right)$. Our goal is to learn a classifier $h_{t}(x)=P\left(Y^{T} \mid X^{T}\right)$ for target domain.

Next, we first define the learning objective of the Transboost as a regularized sum of loss on target domain and weighted source domain. We then purpose our algorithm --- an innovative transfer learning framework of two parallel boosting-tree models that share identical tree structures but different node weights. The special design keeps the merit of robustness and interpretability of tree-based models. Besides, it enables us to train classifier and adjust for distributional differences between domains simultaneously in $O(n)$ complexity, which is much more efficient than traditional kernel method of $O(n\sqrt{n})$ complexity. Finally, we prove that our algorithm is in nature a generalization of the KMM framework \textit{but} using boosting trees as a more flexible kernel method. It also extends KMM to match joint distribution between domains instead of marginal distributions, which potentially improves accuracy in classification problems.

\subsection{Transfer Learning Objective}
We start with a simple learning problem. Given sufficient instances from the target domain, we can directly optimize $h_{t}$ with the following objective:
\begin{equation}
\begin{split}
\mathcal{L}_{T}\left(h_{t}\right) & = \mathbb{E}_{(x, y) \sim P^{T}}\left[\mathcal{L}\left(x, y ; h_{t}\right)\right] \\
& = \frac{1}{n_{T}} \sum_{j}^{n_{T}}\left[\mathcal{L}\left(x_{j}^{T}, y_{j}^{T} ; h_{t}\right)\right],
\end{split}
\end{equation}
where $\mathcal{L}_{T}(h)$ is the expected loss of classifier $h_{t}$ over instances on target domain. $\mathcal{L}$ is cross-entropy loss. 
However, due to the scarcity of instances from the target domain and the curse of high dimensionality, directly optimizing classifier $h_{t}$ often leads to poor prediction performance. To alleviate the issue, transfer learning methods could be used to utilize the information from a source domain as prior knowledge to improve the model performance on the target domain. However, applying the model trained from the source domain may \textit{not} work well because, in many real-world applications, the source domain and target domains are still different. To address that, a common strategy is to assign weights to instances from source domain in the loss function: 
\begin{equation}
\begin{split}
\mathcal{L}_{T}\left(h_{t}\right) &= \mathbb{E}_{(x, y) \sim P^{S}}\left[\frac{P^{T}(x, y)}{P^{S}(x, y)} \mathcal{L}\left(x, y ; h_{t}\right)\right] \\ 
&\approx \frac{1}{n_{s}} \sum_{i}^{n_{S}}\left[ \beta_{i} \mathcal{L}\left(x_{i}^{S}, y_{i}^{S} ; h_{t}\right)\right] \\
& = \mathcal{L}_{S}\left(h_{t}, \beta \right).
\end{split}
\end{equation}
Unlike Eq. (1), the loss is estimated with re-weighted instances from the source domain and the weights $\beta_{i}  = \frac{P^{T}\left(x_{i}^{S}, y_{i}^{S}\right)}{P^{S}\left(x_{i}^{S}, y_{i}^{S}\right)}$ is the ratio between the joint distribution $P^{T}(x, y)$ and $P^{S}(x, y) .$
We then have
$$
\frac{P^{T}(x, y)}{P^{S}(x, y)}=\frac{P^{T}(x)}{P^{S}(x)} * \frac{P^{T}(y \mid x)}{P^{S}(y \mid x)},
$$
where we split the distribution ratio into two parts - marginal distribution ratio $P^{T}(x) / P^{S}(x)$ and conditional distribution ratio $P^{T}(y \mid x) / P^{S}(y \mid x) .$
Later, we will show how to estimate these two parts with boosted tree models.

In practice, the overall objective of TransBoost is defined on both the reweighted source domain and the target domain so that target-specific feature weight can also be learnt during the training process: 
\begin{equation}
\mathcal{L}\left(h_{t}\right)=\lambda\mathcal{L}_{S}\left(h_{t}, \beta \right)+\mathcal{L}_{T}\left(h_{t}\right)+\Omega\left(h_{t}\right),
\end{equation}
where $\beta$ is the weights to match the joint distribution,  $\lambda$ is the hyper-parameter to balance $L_{T}$ and $L_{S}$ , and $\Omega$ is the regularizing term for $h_{t}$.

\subsection{Learning Algorithm}
In this section, we discuss the TransBoost in details. The overall architecture of TransBoost is presented in 
Figure~\ref{tab:Figure 1}. The TransBoost model consists of two parallel GBDTs, a main boosting tree and an ancillary boosting tree, for the target domain and the source domain respetively. The final model is learnt by the main boosting tree which still enjoys effectiveness, interpretability and robustness. At each iteration in the TransBoost, there are two key steps:

\begin{figure}[t]
\centering
    \includegraphics[width=1.0\linewidth]{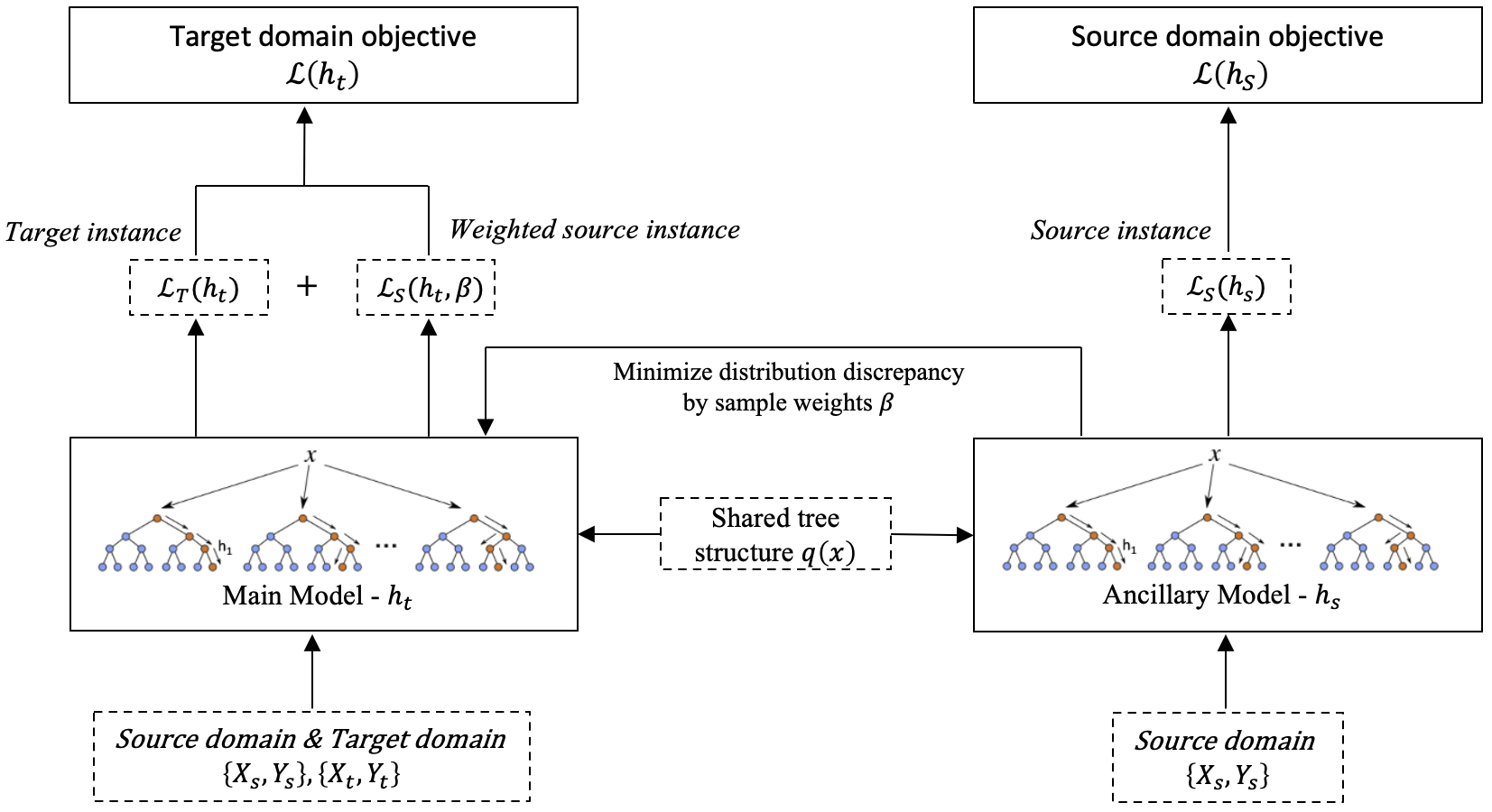}
\caption{The TransBoost learning algorithm.}
\label{tab:Figure 1}
\end{figure}  

\subsubsection{Same Structure Tree Boosting.}

At the \textit{k}-th iteration, the added new tree for the target domain is trained to optimize  $$
\mathcal{L}\left(h_{t}\right)=\lambda\mathcal{L}_{S}\left(h_{t}, \beta^{k} \right)+\mathcal{L}_{T}\left(h_{t}\right)+\Omega\left(h_{t}\right),
$$
while the added new tree for the source domain is trained to optimize $$
\mathcal{L}\left(h_{s}\right)=\mathcal{L}_{S}\left(h_{s}\right)+\Omega\left(h_{s}\right).
$$
However, although the tree trained on the source domain is constraint to have the identical base tree structures and split values as the tree trained for the target domain, it is allowed to have different node weights. At the high level, the identical base tree structures will induce the same partition on both source domain and target domain, which ensures matched marginal distributions in the learning process. On the other hand, different weights provide flexibility to learn different conditional distributions. 

\subsubsection{Efficient Weights Update.}
A major drawback of traditional kernel-trick-based method to reweight source domain is the computational cost of solving the quadratic optimization problem, typically at least in $O(n\sqrt{n})$ time. In our algorithm, the same-structure GBDTs enable to apply boosting-tree as an innovative kernel method. We only use the \textit{k}-th tree to construct the kernel without solving the quadratic optimization problem. Thus an analytical solution to the reweights could be derived in $O(n)$ time. 
More specifically, after adding boosting tree at the \textit{k}-th iteration, we can then update the weights in the objective function as
\begin{equation}
\beta^{k+1}_{i}
=  \frac{n_{I_{q(x_i)}} N_S} {m_{I_{q(x_i)}} N_T}  \frac{y_{i} h_{t}^{k}\left(x_{i}\right)+\left(1-y_{i}\right) \left(1-h_{t}^{k}\left(x_{i}\right)\right)}{y_{i} h_{s}^{k}\left(x_{i}\right)+\left(1-y_{i}\right) \left(1-h_{s}^{k}\left(x_{i}\right)\right)},
\end{equation}
where $I_{q(x_i)}$ as the instance set of leaf that the source domain instance $x_i$ belongs to. $ n_{I_{q(x_i)}}$ is the number of target instances from $I_{q(x_i)}$ and $m_{I_{q(x_i)}}$ is the number of source instances from $I_{q(x_i)}$. $h_t^{k}(x_i)_{s}$ is the estimation of $x_i$ from the main model for the target domain at the k-the iteration while $h_s^{k}(x_i)$ is the estimation of $x_i$ from the ancillary model for the source domain. The TransBoost algorithm proceeds as the \textit{Algorithm 1}.

\begin{figure}[t]
\centering
    \includegraphics[width=0.9\linewidth]{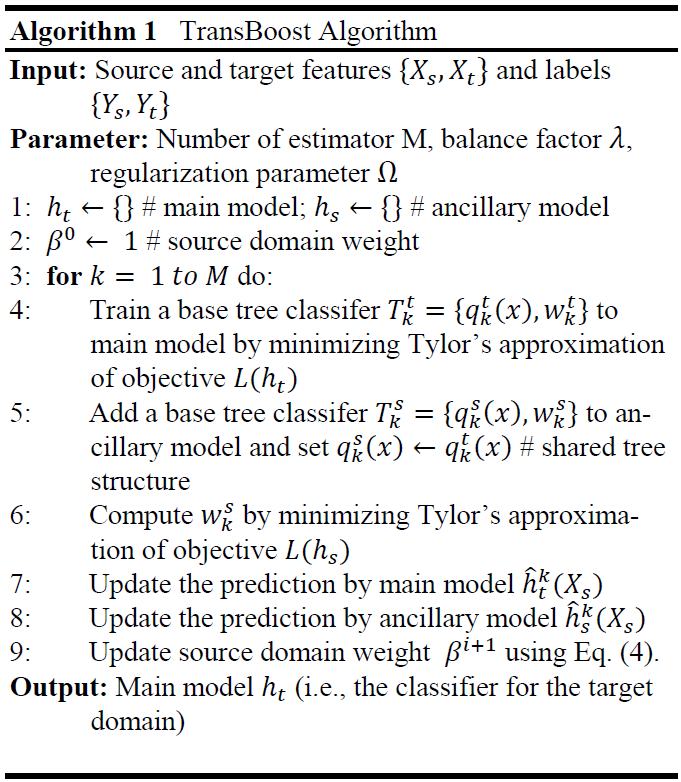}
\label{fig:algorithm}
\end{figure}

\subsection{Theoretical Analysis}
In high-dimensional data,  $\frac{P^{T}(x, y)}{P^{S}(x, y)}$  is difficult to estimate \citep{zhuang2020comprehensive}. KMM \citep{huang2006correcting} has been used to correct sampling bias from marginal distribution for unlabelled data.
In our work, we generalize and customize the KMM framework in three-ways: first, we adopt boosting trees as kernels. Compared to traditional kernels, the kernel is derived from the training data hence more flexible \citep{davies2014erandom, scornet2015random, breiman2000some}; second, we only use the last added tree to construct the tree kernel so that the instance weights could be calculated efficiently; third, for the binary classification problem, we extend KMM to match joint distributions instead of marginal distributions.

For a tree with tree structure function $q(x)$, the tree kernel is defined as the connection function of the tree, $K(x,y) = \mathbbm{1}_{q(x) = q(y) }$.

~\\
\noindent \textbf{Proposition 1} \; \textit{When using tree kernels, The best sampling weights adjustment for the marginal distribution in KMM is $\beta^{marginal}_{i} = \frac{n_{I_{q(x_i)}} N_S} {m_{I_{q(x_i)}}N_{T}}$, where $m_{I_{q(x)}}$ is the number of source instances from $I_{q(x)}$ and $n_{I_{q(x)}}$ is the number of target instances from $I_{q(x)}$, $N_S$ is the total number of samples in the source domain and $N_T$ is the total number of samples in the target domain.}
\begin{proof}
We can first order the nodes in the source domain by their leaves and then $K(x_i,x_j)$ over the source domain will be a block diagonal matrix
$$
K =\left[\begin{array}{cccc}
J_{m_{I_{1}}} & 0 & \cdots & 0 \\
0 & J_{m_{I_{2}} }& \cdots & 0 \\
\vdots & \vdots & \ddots & \vdots  \\
0 & 0 & \cdots & J_{m_{I_{T}}}
\end{array}\right],
$$ where $J_{m_{I_{t}}}$ is a ($m_{I_{t}}$ by $m_{I_{t}}$) matrix of ones. Note that this kernel function is positive semi-definite.
Then we can follow \citep{huang2006correcting} to find the best ratio estimations by optimizing $\beta_{i}$ via  

$$\underset{\beta}{minimize} \;\; \frac{1}{2} \beta^{T} K \beta-k^{T} \beta$$
where $K_{i, j}$ is the above block diagonal matrix and $k_{i}=\frac{N_{S}}{N_{T}} \sum_{j=1}^{n_{T}} K\left(x_{i}, x_{j}^{\prime}\right), N_{s}$ is the number of elements
from the source domain and $N_{t}$ is the number of elements from the target domain, $x_{i}$ is data from the source domain and $x^{\prime}{ }_{j}$ is data from the target domain.

To derive the optimal weights $\beta_i$, let's rearrange the objective function by leaves. Noticing that when $x_i$, $x_j$ are from the same leaf, they have the same weight $\beta_{I_{q(x)}}$, we'll have
\begin{equation}
\frac{1}{2} \beta^{T} K \beta-k^{T} \beta=
\frac{1}{2} \sum_{j=1}^{T} m_{I_{j}} ^{2} \beta_{I_{j}}^{2}- 
\sum_{j=1}^{T}  \frac{m_{I_j} n_{I_j} N_S}{N_T} \beta_{I_j}.
\end{equation}

It's easy to show that these are just $T$ quadratic functions and the optimal weight for leaf $t$ is $\frac{n_{I_{t}} N_S} {m_{I_{t}} N_{T}}$. 
\end{proof}

\noindent \textbf{Proposition 2} \; \textit{For binary classification, the ratio of conditional distribution} $ \frac{P^T(y_i|x_i)}{P^S(y_i|x_i)} $ 
can be approximated by $ \beta^{conditional}_{i} = \frac{y_{i} h_{t}\left(x_{i}\right)+\left(1-y_{i}\right) \left(1-h_{t}\left(x_{i}\right)\right)}{y_{i} h_{s}\left(x_{i}\right)+\left(1-y_{i}\right) \left(1-h s\left(x_{i}\right)\right)}$, 
where $h(x_i) = \\^{p}(y_i=1|x_i)$.

~\\
We train two parallel GBDTs to fit both $h_t(x)$ and $h_s(x)$ and when the fitting is reasonably good, we could use them to approximate the conditional probability. 

With these two propositions, we can proceed handle marginal distributions and conditional distributions.

~\\
\noindent \textbf{Proposition 3} \; \textit{The optimal weights that match the joint distribution is $\beta_i = \beta^{marginal}_i *\beta^{conditional}_i $.}

\begin{proof}
To generalize the KMM approach to matching joint distribution for binary classification problem, we define a new kernel-like function based on the above tree connection function. More specifically, $K_{new}([x_i,y_i],[x_j,y_j]) =  \mathbb{1}_{q(x_i) = q(x_j) } *  \mathbb{1}_{y_i = y_j }$

$$
K_{new} =\left[\begin{array}{ccccc}
J_{m_{I_{1}},1} & 0 & \cdots & 0 & 0 \\
0 & J_{m_{I_{1},0} } & \cdots & 0  & 0\\
\vdots & \vdots & \ddots & \vdots & \vdots \\
0  & 0 &\cdots & J_{m_{I_{T},1}} & 0 \\
0 & 0 & \cdots & 0 & J_{m_{I_{T},0}}
\end{array}\right],$$
where $J_{m_{I_{t}},l}$ is a ($m_{I_{t}} * P^{S}_{I_k,l}$ by $m_{I_{t}} * P^{S}_{I_k,l}$) matrix of ones and $P^{S}_{I_k,l} = P^S(y = l| I_k)$.

Then we can rearrange the objective function by leaves and labels. The objective function takes the following form: 
\begin{multline*}
\frac{1}{2} \beta^{T} K \beta-k^{T} \beta=\frac{1}{2} \sum_{j=1}^{T} \sum_{l=0}^{1}\left(m_{I_{j}} p_{I_{j}, l}^{S}\right)^{2} \beta_{I_{j, l}}^{2}- \\ 
\sum_{j=1}^{T} \sum_{l=0}^{1} \frac{m_{I_{j}} P ^{S}_{I_{j},l} n_{I_{j}} P^{T}_{I_{j},l} N_S }{N_T} {\beta_{j}}_{I_{j}, l},
\end{multline*}
and the analytical solution is
$$
\beta_{i}=\frac{n_{I_{q\left(x_{i}\right)}}{N_{S}}}{m_{I_{q}\left(x_{i}\right)} N_{T}} \frac{p^{T}_{ I_{q\left(x_{i}\right)}, l\left(x_{i}\right)}}{p^{S}_{I_{q\left(x_{i}\right)}, l\left(x_{i}\right)}}.
$$
by proposition 2, we can estimate the ratio of conditional distributions by $ \frac{y_{i} h_{t}\left(x_{i}\right)+\left(1-y_{i}\right) \left(1-h_{t}\left(x_{i}\right)\right)}{y_{i} h_{s}\left(x_{i}\right)+\left(1-y_{i}\right) \left(1-h s\left(x_{i}\right)\right)}$ and the optimal weight is

\begin{equation} \label{eq1}
\begin{split}
\beta_{i} &=\beta^{marginal}_i *\beta^{conditional}_i\\
&=\frac{n_{I_{q(x_i)}} N_S} {m_{I_{q(x_i)}} N_T}  \frac{y_{i} h_{t}\left(x_{i}\right)+\left(1-y_{i}\right) \left(1-h_{t}\left(x_{i}\right)\right)}{y_{i} h_{s}\left(x_{i}\right)+\left(1-y_{i}\right) \left(1-h s\left(x_{i}\right)\right)}.
\end{split}
\end{equation}
\end{proof}

\subsubsection{Algorithm Limitations.}
Finally, because the TransBoost is a \textit{domain-knowledge-independent} algorithm and the algorithm is \textit{only} based on local smoothness assumption, there would be cases when the framework may fail. We elaborate on three different situations that can be addressed in future studies:

\begin{itemize}
     \item \textit{When data is generated from the composition of features}: In this case, there could be non-local features shared between source domain and target domain (e.g., spatial features in image and temporal features in text), and the boosting-tree framework may not work well.
    
     \item \textit{When domain knowledge plays an important role}: from the regularization perspective, domain knowledge could be modeled as a special regularization term. Because our algorithm only applies local smoothness regularization, in this case, it may have sub-optimal performance.
    
    \item \textit{When source and target domains are irrelevant}: in this case, the conditional weight update, $ \frac{y_{i} h_{t}\left(x_{i}\right)+\left(1-y_{i}\right) \left(1-h_{t}\left(x_{i}\right)\right)}{y_{i} h_{s}\left(x_{i}\right)+\left(1-y_{i}\right) \left(1-h s\left(x_{i}\right)\right)}$, will quickly detect that the tree-structure from source domain is less relevant and thus increase the weights of the target domain. As the weights of the source domain decay, the performance of the algorithm will be very comparable to the model that only uses target domain as input.
\end{itemize}

\section{Experiments}
\subsection{Datasets}
We adopt three datasets to evaluate the TransBoost against several baseline methods. The \textit{LendingClub} and \textit{Wine Quality} are public datasets with small sample sizes. They are used to compare the TransBoost with the traditional kernel-related methods that have high computational overhead characteristics. The \textit{Tencent Mobile Payment} dataset is a private large-scale dataset used to showcase the application of our method and compare model performance in the real-world industrial environment.

\subsubsection{Tencent Mobile Payment Dataset.}
We obtain a unique real-world financial fraudulent detection dataset from the Tencent Mobile Payment.\footnote{The dataset in this paper is properly sampled only for testing purpose and does not imply any commercial information. All private information of users is removed from the dataset. Besides, the experiment was conducted locally on Tencent’s server by formal employees who strictly followed data protection regulations.} The dataset is sampled from two business scenarios --- a mature financial product (source domain) with 1.49 million users and a newly-launched product(target domain) with 85,133 users. Each user has 2,714 features and is labeled as fraudulent (i.e., positive) or not (i.e., negative). We randomly choose 44.4\% of users from the target domain for testing. The data description is shown in Table \ref{tab:introduction_tencent}.  


\begin{table}[h]
\centering
\begin{tabular}{lccccccc}
\hline
 \text { Domain } & \text { \#Total } & \text { Positive Rate } \\
 \hline\hline
  \text { Source for Training} & 1,487,904 & 4.99 \% \\
\text { Target for Training} & 49,669 & 4.85 \% \\
\text { Target for Testing} & 39,735  & 4.69 \% \\

\hline
    \end{tabular}
    \caption{Description of Tencent mobile payment dataset.}
    \label{tab:introduction_tencent}
\end{table}


\subsubsection{LendingClub Dataset.}
LendingClub dataset is an open-source micro-lending dataset from LendingClub \cite{lendingclub}. The dataset includes loans for different purposes in many years. To increase the distribution difference, we set the medical loans in 2015 as the source domain and the car loans in 2016Q1 as the target domain. The features include 110 borrower and loan characteristics such as borrower's income, loan's interest rate, etc. The labels are given by loans' final status: fully paid or charged off. We use 5,080 labeled source instances and 1,000 labeled target instances for training, and 3,000 target instances for testing.

\subsubsection{Wine Quality Dataset.}
Wine quality dataset is a popular dataset from UCI Machine Learning Repository \cite{dua2019uci} with two datasets related to red and white wine samples. We set the red wine as the source domain and white wine as the target domain. There are 11 physical and chemical features of wine in the dataset, and the binary outcome label is whether the quality score is higher than 5. We use 3,918 labeled source instances and 500 labeled target instances for training, and 960 target instances for testing.

\text

\subsection{Algorithms for Comparison}
We compared our proposed TransBoost with two types of state-of-the-art transfer learning methods: the traditional and deep transfer learning methods. 

\begin{itemize}
\item Traditional transfer learning methods:
\end{itemize}

\textbf{[1] KMM:} KMM \cite{huang2006correcting} estimates the source instances weight by matching the means in RKHS to adapt marginal distribution. 

\textbf{[2] TrAdaBoost:} TrAdaBoost \cite{dai2007boosting} extends AdaBoost to assign larger training weight to same distribution samples.

\textbf{[3] JDA:} JDA \cite{long2013transfer} minimizes both the marginal and the conditional distribution difference in a low-dimensional feature space with fixed balance factor.

\textbf{[4] CORAL:} CORAL \cite{sun2016return} aligns the second-order statistic features by constructing the transformation matrix.

\textbf{[5] BDA:} BDA \cite{wang2017balanced} minimizes both the marginal and the conditional distribution difference in a low-dimensional feature space with dynamic balance factor.

\begin{itemize}
\item Deep transfer learning methods:
\end{itemize}

\textbf{[6] DAAN:} DAAN \cite{yu2019transfer} is a deep adversarial network that dynamically learns the domain-invariant representations by evaluating the importance of global and local distribution.

\textbf{[7] DDAN:} DDAN \cite{wang2020transfer}, also named as DeepMEDA, is a deep network that quantitatively evaluates the importance of marginal and conditional distribution.

\textbf{[8] DSAN:} DSAN \cite{zhu2020deep} is a deep transfer network that aligns the relevant subdomain distributions based on local maximum mean discrepancy.



\subsection{Implementation Details}
We implement the TransBoost algorithm based on the source code of XGBoost. For fair comparison, we also take XGBoost as the base classifier for traditional benchmarks [1---5]. In the LendingClub and Wine quality datasets, we set the number of estimators (i.e., booster) to 40 and tree depth to 4 for TransBoost and benchmark algorithms. In the Tencent mobile payment dataset, we set the number of estimator = 400 and tree depth = 4. We use standard one-layer Multi-layer perceptron in PyTorch for deep transfer learning benchmarks [6---8] , which is sufficient to solve the classification problems in our experiments \cite{shen2018wassetein}. The learning rate is set to 0.01 and the number of epochs is set to 10.  

Considering that some benchmarks [3, 5, 6, 7, 8] are designed to solve unsupervised domain adaptation problems, for fair comparison, we fine tune them to make full use of the labeled target domain instances. Specifically, we replace the pseudo label with true label in the domain adaption process and add the training of labeled target instances.

All experiments are conducted on the Tencent's server with 64-core CPU and 128G memory. We compare all eight benchmarks on the three datasets.
Besides, we conduct sensitivity analysis to validate that TransBoost can outperform benchmark algorithms with different sizes of the target domain. To simulate the growth of newly launched product, the models are trained with all source domain samples and different fractions of the target domain samples. We use a fix-sized testing dataset to evaluate the model performance with AUC. 

\subsection{Results} 
\subsubsection{Prediction Accuracy.}
\begin{table*}[t]
    \centering
    \begin{tabular}{lcccccccccc}
    \hline

&10\%&20\%&30\%&40\%&50\%&60\%&70\%&80\%&90\%&100\%\\
    \hline
    \hline
Transboost&\textbf{0.7209}&\textbf{0.7253}&\textbf{0.7229}&\textbf{0.7238}&\textbf{0.7227}&\textbf{0.7276}&\textbf{0.7223}&\textbf{0.7213}&\textbf{0.7245}&\textbf{0.7204}\\
Tradaboost&0.7185&0.7059&0.7136&0.7047&0.7081&0.7164&0.7075&0.7068&0.7069&0.7071\\
BDA&0.6899&0.7035&0.7015&0.6908&0.7076&0.7016&0.6971&0.7058&0.7008&0.7041\\
JDA&0.7069&0.6913&0.6994&0.6889&0.7032&0.6979&0.7073&0.7087&0.7136&0.7061\\
CORAL&0.7107&0.7176&0.7089&0.7136&0.7066&0.7131&0.7152&0.7154&0.7154&0.7181\\
KMM&0.6995&0.7019&0.7044&0.7081&0.7077&0.7089&0.7147&0.7205&0.7195&0.7190\\
DSAN&\multicolumn{1}{l}{-}&\multicolumn{1}{l}{-}&\multicolumn{1}{l}{-}&0.6793&0.6978&0.7086&0.6931&0.6954&0.7001&0.6997\\
DAAN&\multicolumn{1}{l}{-}&\multicolumn{1}{l}{-}&\multicolumn{1}{l}{-}&0.6744&0.6909&0.7003&0.6976&0.7067&0.7074&0.7077\\
DeepMEDA&\multicolumn{1}{l}{-}&\multicolumn{1}{l}{-}&\multicolumn{1}{l}{-}&0.6745&0.7016&0.6999&0.7015&0.7014&0.7061&0.7084 \\
\hline

    \end{tabular}
    \caption{AUCs on LendingClub dataset output by different algorithms given a sample fraction of target domain for training.}
    \label{tab:results_lc}
\end{table*}

\begin{table*}[t]
    \centering
    \begin{tabular}{lcccccccccc}
    \hline
&10\%&20\%&30\%&40\%&50\%&60\%&70\%&80\%&90\%&100\%\\
\hline\hline
Transboost&0.7821&\textbf{0.8114}&\textbf{0.8117}&\textbf{0.8268}&\textbf{0.8224}&\textbf{0.8225}&\textbf{0.8261}&\textbf{0.8254}&\textbf{0.8279}&0.8290\\
Tradaboost&0.7776&0.8023&0.8008&0.8039&0.8156&0.8093&0.8208&0.8240&0.8211&\textbf{0.8325}\\
BDA&\textbf{0.8080}&0.7873&0.7888&0.8117&0.8151&0.8005&0.8085&0.8187&0.8257&0.8191\\
JDA&0.7898&0.7914&0.7987&0.7977&0.8015&0.7993&0.8024&0.8003&0.8080&0.8112\\
CORAL&0.7777&0.7916&0.8105&0.8072&0.8119&0.8065&0.8158&0.8174&0.8207&0.8215\\
KMM&0.7765&0.7989&0.8013&0.7878&0.7918&0.8022&0.7962&0.7971&0.8052&0.8067\\
DSAN&0.7484&0.7892&0.8097&0.8103&0.8139&0.8092&0.8101&0.8123&0.8148&0.8133\\
DAAN&0.7724&0.7961&0.8112&0.8066&0.8008&0.8015&0.8008&0.8018&0.8053&0.8073\\
DeepMEDA&0.7661&0.7897&0.8117&0.8114&0.8117&0.8111&0.8093&0.8115&0.8127&0.8096\\
\hline
    \end{tabular}
    \caption{AUCs on Wine quality dataset output by different algorithms given a sample fraction of target domain for training.}
    \label{tab:results_wine}
\end{table*}

\begin{figure}[t]
\centering
    \includegraphics[width=0.9\linewidth]{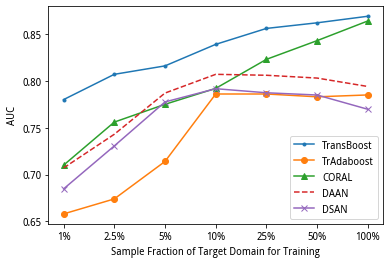}
\caption{AUCs on Tencent mobile payment dataset output by different algorithms given a sample fraction of target domain for training.}
\label{fig:relative_auc}
\end{figure}

The AUCs of TransBoost and other baseline methods on LendingClub, Wine quality, and Tencent mobile payment datasets are shown in Tables \ref{tab:results_lc}, \ref{tab:results_wine}, and Figure \ref{fig:relative_auc}, 
respectively. Note that fewer results are shown in Figure \ref{fig:relative_auc} because some of the traditional benchmark models ([1, 3, 5, 7]) cannot handle the large volume and high sparsity of the Tencent mobile payment dataset. This further shows the scalability of our algorithm. All methods can achieve better prediction results with a larger size of the target domain. The TransBoost achieves rather stable prediction results even with small-sized target domain data. The TransBoost outperforms other baselines in almost all cases, showing its superiority in prediction.

\subsubsection{Model Runtime.}
\begin{figure}[t]
\centering
    \includegraphics[width=0.9\linewidth]{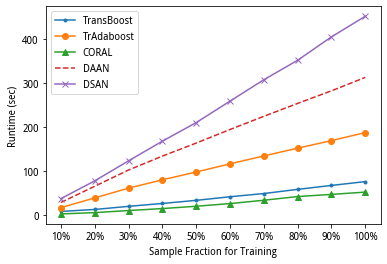}
\caption{Runtime of different algorithms on Tencent mobile payment dataset with different training sample sizes.}
\label{fig:runtime}
\end{figure}

To further compare the efficiency of different models, we run each model 10 times with different fractions of the Tencent mobile payment dataset, and calculate the average runtime as shown in Figure \ref{fig:runtime} (again, several benchmarks fail to converge on this large-scale dataset). Overall, the TransBoost presents superior efficiency compared with most baseline models such as DSAN and DAAN. The runtime grows linearly to sample size. Considering both efficiency and prediction accuracy, we conclude that the TranBoost achieves high prediction accuracy without sacrificing efficiency, which fits perfectly for industrial application. 

\subsubsection{Robustness to Data Sparsity.}

\begin{figure}[t]
\centering
    \includegraphics[width=0.9\linewidth]{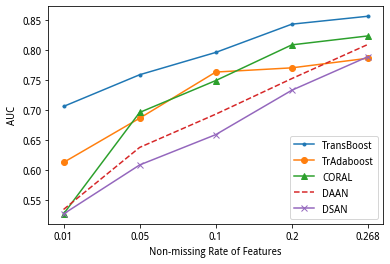}
\caption{AUCs of different models under diverse data sparsity levels on Tencent mobile payment dataset.}
\label{fig:sparsity}
\end{figure}  

The Tencent mobile payment dataset is sparse in nature with an overall non-missing rate of 26.8\%. To validate that our algorithm is robust in more sparse contexts, we further lower the non-missing rate and compare the model performance. Specifically, we randomly set more features to \textit{NULL} value in the training samples of Tencent dataset to simulate datasets with varied sparsity of [1\% to 25\%] and test the TransBoost against other baselines. All the experiments are carried out 10 times and the average performance is shown in Figure \ref{fig:sparsity}. TransBoost is robust enough and outperform all the baselines at various sparsity levels. 




\subsubsection{Model Interpretation.}

\begin{figure}[t]
\centering
    \includegraphics[width=1.0\linewidth]{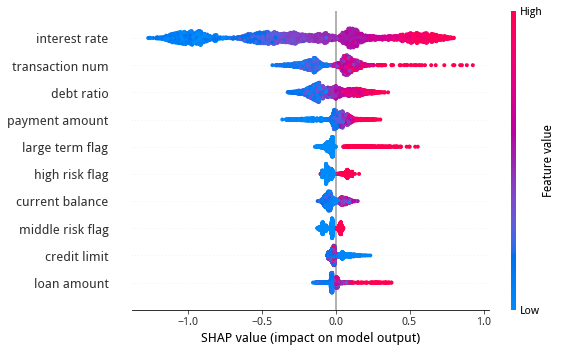}
\caption{SHAP Value of top 10 important features for predicting financial risk output TransBoost model based on LendingClub dataset.}
\label{tab:SHAP value}
\end{figure}

The tree-based TransBoost has the natural advantage in model interpretation. Figure \ref{tab:SHAP value}
presents the Shapley value of important features \citep{lundberg2017unified} for financial risk evaluation based on the LendingClub dataset. The figure offers meaningful insights by indicating that loan interest rate, transaction number, debt ratio, and payment amount, are significant for predicting loan default behavior. These findings are consistent with previous studies (e.g., \citet{lu2020profit}). 

\subsubsection{Financial Inclusion.}

\begin{table}[t]
    \centering
    \begin{tabular}{lcccccccccc}
    \hline
&\multicolumn{1}{r}{8\%}&\multicolumn{1}{r}{9\%}&\multicolumn{1}{r}{10\%}&\multicolumn{1}{r}{11\%}&\multicolumn{1}{r}{12\%}\\
\hline\hline
TransBoost&\textbf{0.5423}&\textbf{0.6333}&\textbf{0.7477}&\textbf{0.8010}&\textbf{0.8567}\\
TrAdaBoost&0.4917&0.5607&0.6543&0.7523&0.8263\\
BDA&0.5043&0.6123&0.6613&0.7607&0.8280\\
JDA&0.5170&0.6210&0.6913&0.7617&0.8140\\
CORAL&0.4920&0.5933&0.6790&0.7640&0.8120\\
KMM&0.5403&0.6260&0.7087&0.7823&0.8460\\
DSAN&0.4513&0.5390&0.6040&0.6767&0.7457\\
DAAN&0.4047&0.5460&0.6560&0.7460&0.7950\\
DeepMEDA&0.4963&0.5833&0.6493&0.7037&0.7983\\
\hline
    \end{tabular}
    \caption{Loan approval ratios under different algorithms given a default rate.}
    \label{tab:results_number}
\end{table}

\begin{table}[t]
    \centering
    \begin{tabular}{lcccc}
    \hline
&\textbf{8\%}&&\textbf{10\%}&\\
&Income&Rent ratio&Income&Rent ratio\\
\hline \hline
TransBoost&76,257&\textbf{0.4336}&\textbf{72,902}&\textbf{0.4505}\\
TrAdaBoost&76,284&0.4315&73,780&0.4465\\
BDA&\textbf{74,254}&0.3908&72,988&0.4446\\
JDA&76,717&0.4161&73,212&0.4457\\
CORAL&76,370&0.4108&74,437&0.4445\\
KMM&77,137&0.4235&72,986&0.4419\\
\hline
    \end{tabular}
    \caption{Socioeconomic features of the approved users by different algorithms at default rate of 8\% and 10\% (unit of Income: USD).}
    \label{tab:results_level}
\end{table}

Finally, to examine whether the TransBoost can better advance finance inclusion than do other algorithms. Based on the LendingClub dataset,\footnote{We checked on all the three datasets and obtained consistent findings. Due to the Tencent data protection regulations, we do not report the results on its dataset.} we calculate the loan approval ratios under different algorithms given a default rate from 8\% to 12\%. Table \ref{tab:results_number} reveals that at all the given financial risk levels, the TransBoost yields highest loan approval ratio; that is, the TransBoost enables financial service providers to serve the largest number of users.

We further analyze two typical user features of the approved users. In specific, we calculate the average annual income and ratio of house renting for the approved users given the default rate of 8\% and 10\%. Table \ref{tab:results_level} indicates that generally, the users selected by TransBoost have lower income and higher proportion of house renting than do the users selected by other algorithms. It suggests that compared with other algorithms, the TransBoost not only offers financial service access to broader users but also covers users who would otherwise be excluded by other algorithms due to their less desirable background. Therefore, the TransBoost improves financial inclusion.

\section{Conclusion}
Achieving financial inclusion is crucial for mitigating financial inequality. However, reaching out to broader users incurs new challenges 
to financial risk evaluation. This paper develops a novel TransBoost algorithm which combinines the merits of tree-based models and kernel methods. The experiments on three datasets show that the TransBoost outperforms all baselines in prediction accuracy and efficiency, shows stronger robustness to data sparsity, and provides clear model interpretation. Moreover, we demonstrate the value of the TransBoost in improving financial inclusion.


\section{Acknowledgement}
T. Lu, C. Wang, and Y. Xu thank the National Natural Science Foundation of China of Grants No. 71872050, 72101007, and 71531006, respectively.

\newpage

\bibliography{aaai22.bib}

\end{document}